\documentclass{article}
\usepackage{spconf,amsmath,graphicx,hyperref}
\usepackage{multirow}
\usepackage{booktabs}
\usepackage{algorithm}
\usepackage{algorithmic}
\usepackage{makecell}
\usepackage{amssymb}

\title{SAGE: Semantic-Aware Shared Sampling for Efficient Diffusion}

\name{Haoran Zhao \quad Tong Bai$^{*}$ \thanks{$^*$Tong Bai is the corresponding author.} \quad Lei Huang \quad Xiaoyu Liang}
\address{Beihang University, Beijing, China\\
Email: \{zhr23, tongbai, huangleiai, xiaoyuliang\}@buaa.edu.cn}
\begin{document}
\maketitle
\begin{abstract}
Diffusion models manifest evident benefits across diverse domains, yet their high sampling cost, requiring dozens of sequential model evaluations, remains a major limitation. 
Prior efforts mainly accelerate sampling via optimized solvers or distillation, which treat each query independently. In contrast, we reduce total number of steps by sharing early-stage sampling across semantically similar queries.
To enable such efficiency gains without sacrificing quality, we propose SAGE, a semantic-aware shared sampling framework that integrates a shared sampling scheme for efficiency and a tailored training strategy for quality preservation.
Extensive experiments show that SAGE reduces sampling cost by 25.5\%, while improving generation quality with 5.0\% lower FID, 5.4\% higher CLIP, and 160\% higher diversity over baselines.
\end{abstract}
\section{Introduction}
\label{sec:intro}

Diffusion models are a powerful class of generative models that iteratively recover data distributions from Gaussian noise.
They have demonstrated remarkable capabilities across diverse domains, including image synthesis \cite{dhariwal2021diffusion,esser2024scaling}, video generation \cite{sora,10888199}, molecule design \cite{watson2023novo}, and decision-making \cite{janner2022planning}.

However, a key limitation of diffusion models lies in their high computational cost of sampling. 
The generation process can be formulated as solving stochastic or ordinary differential equations (SDEs/ODEs) with time-dependent score functions estimated by neural networks \cite{song2020score}. In practice, the continuous trajectories are discretized into tens or hundreds of steps, each requiring a network evaluation.
Combined with the large size of modern diffusion models, this leads to extremely costly inference.
To address these challenges, prior works have explored more efficient ODE solvers to reduce the number of iterations \cite{song2020denoising,lu2022dpm} and distillation techniques that compress the process into fewer-step student models \cite{meng2023distillation,luo2023latent}.

\begin{figure}[t!]\center
\includegraphics[width= 0.47 \textwidth]{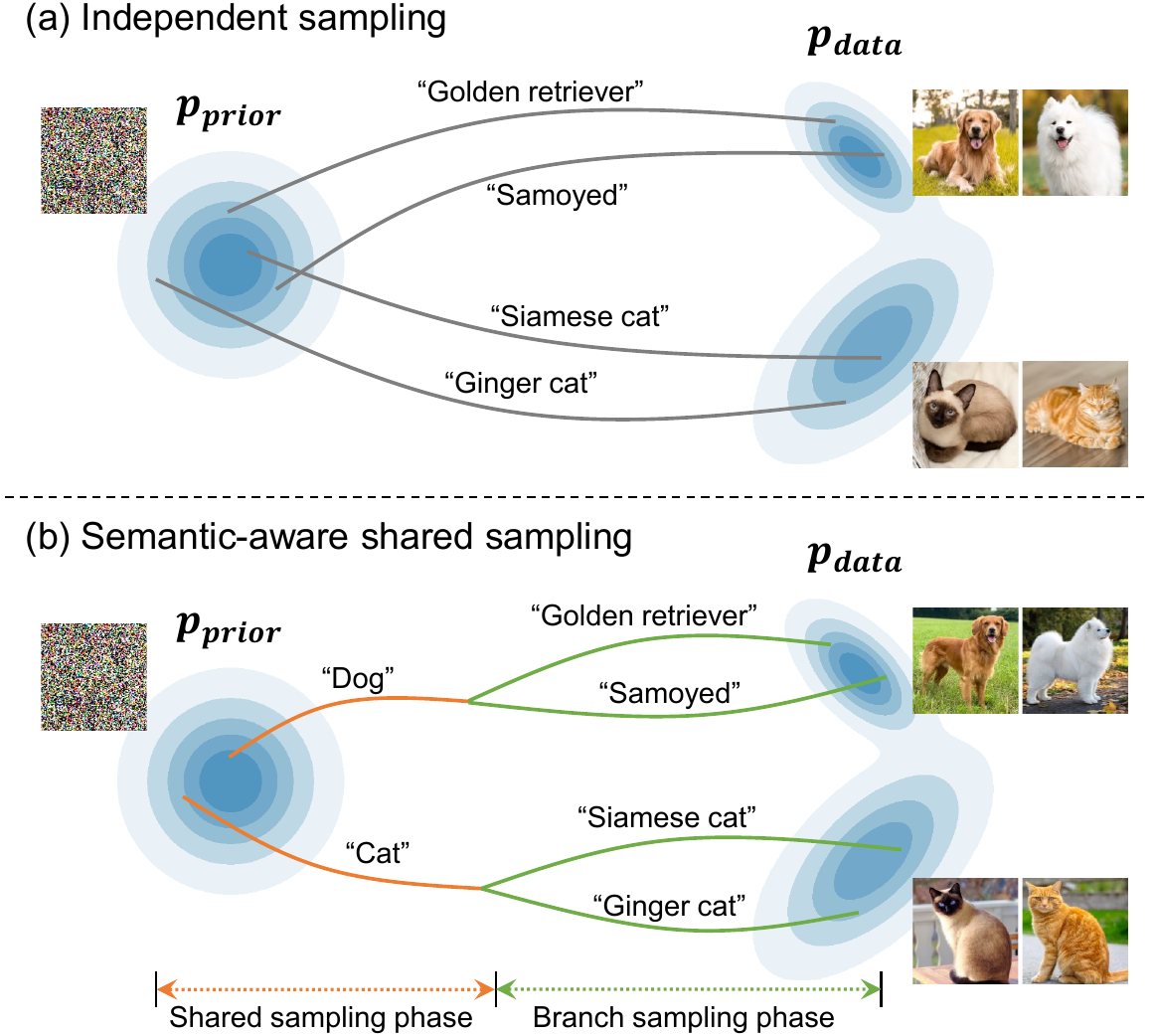}
\vspace{-2pt}
\caption{(a) Conventional independent sampling for diffusion, and (b) semantic-aware shared sampling, where semantically similar prompts share early steps to reduce resource cost.}
\vspace{-4pt}
\label{fig:architecture}
\end{figure}

In this paper, we explore a new direction for improving the trade-off between generation quality and efficiency---\textit{sharing early sampling steps across semantically similar queries}. 
As illustrated in Fig.~\ref{fig:architecture}, conventional independent sampling processes each prompt separately, whereas in our shared sampling strategy, prompts are first grouped by semantics for shared sampling (e.g., dog), then refined in a branch phase with prompt-specific guidance (e.g., golden retriver, samoyed). This design reduces the total number of sampling steps by reusing early computations across prompts. The key challenge, however, lies in achieving such efficiency gains without compromising prompt-specific generation quality.

Recent work has explored shared sampling in edge-device collaborative computing scenario, where edge server executes shared sampling steps and end devices receive intermediate results from server to perform local sampling \cite{11059888}. This distributed design reduces resource cost and improves scalability.
Yet, two important issues remain open: (i) existing methods do not address how to group prompts by semantic similarity or construct effective shared guidance; and (ii) they rely on pre-trained diffusion models without adaptation, which limits model’s ability to accommodate the shared sampling scheme and ultimately constrains the achievable quality-efficiency trade-off.
To fill these gaps, we propose  \textbf{SAGE}, \underline{S}emantic-\underline{A}ware shared samplin\underline{G} for \underline{E}fficient diffusion, a framework that integrates a shared diffusion sampling scheme and a tailored shared diffusion training strategy.

Our contributions are as follows.
\vspace{-8pt}
\begin{itemize}\setlength\itemsep{0pt}
    \item \textit{Semantic-aware shared diffusion sampling}: We design a sampling scheme that groups prompts by semantic similarity and performs shared sampling followed by branch sampling within each group, effectively reducing the total number of steps.
    
    \item \textit{Soft-target guided shared diffusion training}: We propose a fine-tuning strategy with a novel loss $\mathcal{L}_{\text{SAGE}}$ that integrates soft-target alignment and dual-phase consistency regularization, enabling the model to preserve both group-level semantic coherence and prompt-specific fidelity under shared sampling.
    
    \item \textit{Experimental evaluation}: We construct a semantically grouped dataset from MS COCO 2017 and conduct extensive experiments. Results show that SAGE consistently outperforms baseline models, achieving 5.0\% lower FID, 5.4\% higher CLIP score, and 160.0\% higher diversity when 40\% of steps are shared.
\end{itemize}

\section{Methods}
\label{sec:methods}

We first revisit standard diffusion formulations. 
Building on this, we introduce SAGE, which integrates (i) a semantic-aware shared sampling scheme to reduce sampling steps, and (ii) a tailored training strategy to maintain generation quality.

\subsection{Text-to-Image Diffusion Models}

We follow the standard latent diffusion model (LDM) framework, where the forward process gradually perturbs a latent sample $\mathbf{z}$ by adding Gaussian noise \cite{karras2022elucidating}:
\vspace{-3pt}
\begin{equation}
q_t(\mathbf{z}_t\mid\mathbf{z}_0)=\mathcal N(\alpha_t\mathbf{z}_0,\;\sigma_t^2\mathbf I),
\end{equation}
with $\alpha_t,\sigma_t$ denote the noise schedule. 
The reverse process is parameterized by a neural network $\epsilon_\theta(\cdot)$, which learns to predict the noise $\boldsymbol\epsilon$ injected at step $t$, conditioned on text prompt $\mathbf{c}$. The training objective is typically the mean squared error (MSE) between the predicted and true noise \cite{ho2020denoising}:
\vspace{-3pt}
\begin{equation}
\mathcal{L}_{\text{LDM}}=
\mathbb{E}_{\mathbf{z}, \mathbf{c}, \boldsymbol\epsilon,t} \left[ w_t ||\epsilon_\theta(\alpha_t \mathbf{z} + \sigma_t\boldsymbol{\epsilon}, \mathbf{c})-\boldsymbol{\epsilon}||^2 \right],
\end{equation}
where $w_t$ is a time-dependent weight. At inference, the reverse process is typically approximated with numerical SDE/ODE solvers, progressively refining Gaussian noise into structured latents that are then decoded into pixel space.

\begin{algorithm}[t]
\caption{Shared Diffusion Sampling}
\label{alg:sampling}
\begin{algorithmic}[1]
\REQUIRE Prompts $\{\mathbf{p}^m\}_{m=1}^M $, diffusion model $\epsilon_\theta$, text encoder $\mathcal{T}$, latent decoder $\mathcal{D}$, branch point $T^*$
\STATE Encode text prompts $\mathbf{c}^m = \mathcal{T}(\mathbf{p}^m)$
\STATE Partition prompts into $K$ groups $\{\mathcal{G}_k\}_{k=1}^K$
\FOR{each group $\mathcal{G}_k = \{\mathbf{p}^n\}_{n=1}^N$}
 \STATE Sample a shared noise $\mathbf{z}_T \sim \mathcal{N}(0, \mathbf{I})$
 \STATE Compute shared condition $\overline{\mathbf{c}} = \frac{1}{N} \sum_{n=1}^N \mathbf{c}^n$
\FOR[Shared sampling phase]{$t=T, \dots, T^*$} 
 \STATE $\mathbf{z}_{t-1} = \text{sampler.step}(\mathbf{z}_t, t, \overline{\mathbf{c}},\omega(t))$ 
\ENDFOR
\FOR[Branch sampling phase]{$n=1, \dots, N$}
 \STATE $\mathbf{z}_{T^*}^n = \mathbf{z}_{T^*} $
 \FOR{$t=T^*, \dots, 1$}
  \STATE $\mathbf{z}_{t-1}^n = \text{sampler.step}(\mathbf{z}_t^n, t, \mathbf{c}^n,\omega(t))$
 \ENDFOR
 \STATE $\mathbf{x}_0^n \gets \mathcal{D}(\mathbf{z}_0^n)$
\ENDFOR
\ENDFOR
\RETURN $\{\mathbf{x}_0^m\}_{m=1}^{M}$
\end{algorithmic}
\end{algorithm}

\begin{algorithm}[t]
\caption{Shared Diffusion Training}
\label{alg:training}
\begin{algorithmic}[1]
\REQUIRE Grouped dataset $\mathcal{G}$, pre-trained text encoder $\mathcal{T}$, latent encoder $\mathcal{E}$, diffusion model $\epsilon_\theta$, branch point $T^*$
\REPEAT
 \STATE Sample a training group $\{(\mathbf{x}^n, \mathbf{p}^n)\}_{n=1}^{N} \sim \mathcal{G}$
 \STATE Encode images $\mathbf{z}^n = \mathcal{E}(\mathbf{x}^n)$
 \STATE Encode text prompts $\mathbf{c}^n = \mathcal{T}(\mathbf{p}^n)\ $
 \STATE Compute shared rep. $\overline{\mathbf{z}} \!=\! \frac{1}{N} \!\sum_{n=1}^N\! \mathbf{z}^n, \overline{\mathbf{c}} \!=\! \frac{1}{N} \!\sum_{n=1}^N\! \mathbf{c}^n\ $
 \STATE Sample timesteps $t_s \!\sim\! \mathcal U\!\{T^*\!, \!\dots\!, \!T\}\!,\! \ t_b \!\sim\! \mathcal U\!\{1, \!\dots\!, \!T^*\!\}$
 \STATE Sample a shared noise $\boldsymbol{\epsilon} \sim \mathcal{N}(0, \mathbf{I})$
 \STATE Take gradient descent step on $\nabla_\theta \mathcal{L}_{\text{SAGE}}(\theta)$
\UNTIL{converged}
\end{algorithmic}
\end{algorithm}

\subsection{SAGE: Shared Diffusion Sampling}

We propose a shared diffusion sampling algorithm that exploits semantic similarity among prompts to reduce total sampling steps for a batch $\{\mathbf{p}^m\}_{m=1}^M$. 

First, the batch is partitioned into groups based on semantic similarity. Since text-to-image diffusion models typically employ a text encoder to generate embeddings for input prompts, we can reuse these embeddings to measure similarity. Specifically, prompts are grouped such that for each group $\mathcal{G}_k$, the cosine similarity between embeddings of any two prompts exceeds a predefined threshold.

Image generation is then performed group-wise. For each group, only one shared initial noise is sampled. In the \textit{shared phase}, the model is conditioned on the average of all text embeddings in the group, which guides the common sampling trajectory. In the subsequent \textit{branch phase}, each prompt continues sampling from the shared intermediate state, guided by its own text embedding. The branch point $T^*$ can be fixed or adaptively chosen based on prompt similarity.
The full procedure is summarized in Algorithm~\ref{alg:sampling}.

\begin{figure*}[t!]\center
\includegraphics[width= 0.98 \textwidth]{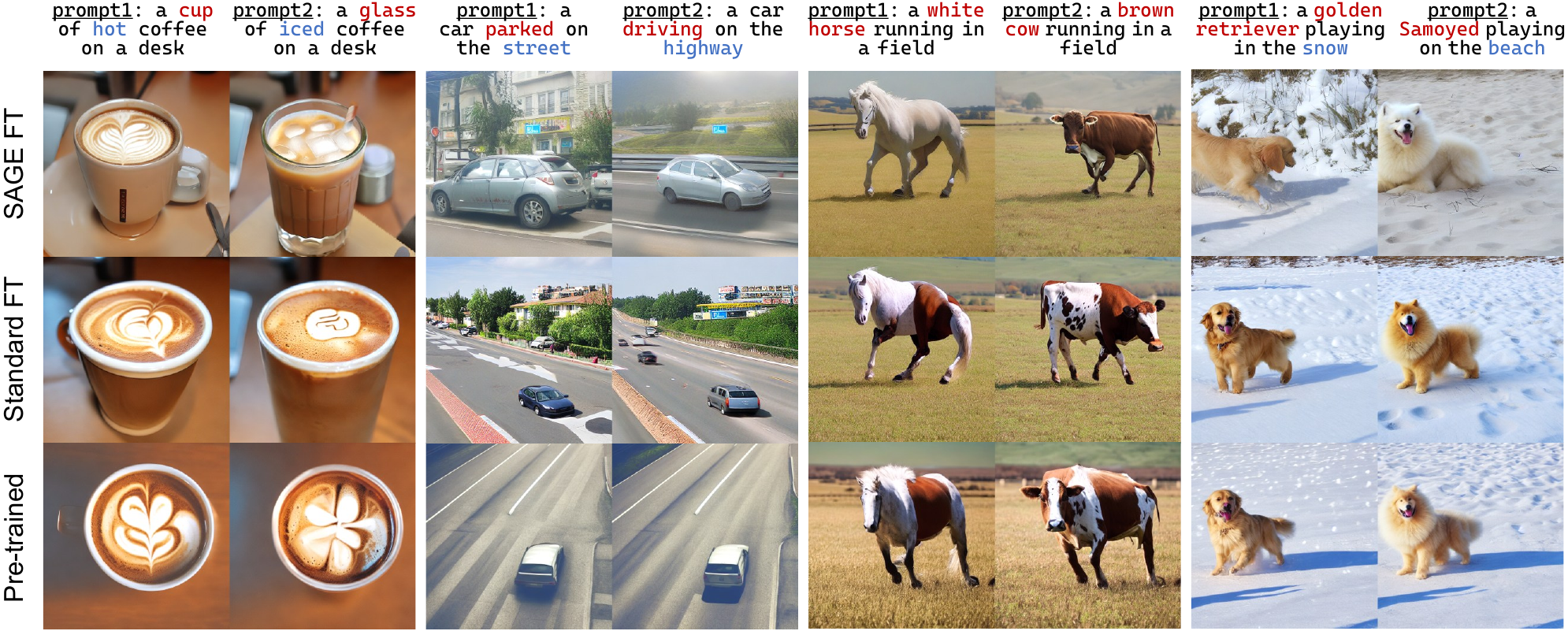}
\vspace{-10pt}
\caption{Examples of generated images under the shared sampling scheme across three methods: pre-trained Stable Diffusion v1.5 (Pre-trained), standard fine-tuning (Standard FT), and the proposed SAGE fine-tuning (SAGE FT). All results are obtained using DDIM \cite{song2020denoising} with 30 total steps, where 9 steps are shared across prompts.}
\vspace{-3pt}
\label{fig:images}
\end{figure*}

\subsection{SAGE: Shared Diffusion Training}

In this paper, we also propose a training algorithm and design a tailored objective $\mathcal{L}_{\text{SAGE}}$ that can improve generation quality under the shared sampling scheme. The overall procedure is summarized in Algorithm~\ref{alg:training}.

Prompts and images are first partitioned into groups according to prompt semantic similarity, forming a grouped dataset $\mathcal{G}$ used to fine-tune the diffusion model $\epsilon_\theta$.
For a group of semantically similar pairs $\{(\mathbf{x}^1, \mathbf{p}^1),\cdots,(\mathbf{x}^N, \mathbf{p}^N)\}$, we encode both images and prompts into latent space and compute shared representations $\overline{\mathbf{z}}$ and $\overline{\mathbf{c}}$.
The key idea is to align the early denoising trajectory with the group-level semantic average, while preserving prompt-specific fidelity in later stages. To this end, we sample two timesteps in each training iteration: $t_s$ for the \textit{shared phase} and $t_b$ for the \textit{branch phase}.

We design a hybrid objective that jointly optimizes both shared and branch phases:
\vspace{-5pt}
\setlength{\jot}{1pt}
\begin{equation}
\begin{aligned}
\mathcal{L}_{\text{SAGE}}=&
\underset{\mathbf{z}, \mathbf{c}, \boldsymbol\epsilon,t_{s},t_{b}}{\mathbb{E}} \! \Bigg[ \!
\ \lambda_{1}\,w_{t_s}\Bigr\|\epsilon_\theta(\alpha_{t_s} \! \overline{\mathbf{z}} \!+\! \sigma_{t_s} \! \boldsymbol{\epsilon},\overline{\mathbf{c}})-\boldsymbol{\epsilon}\Bigr\|^{2}\\
&+\lambda_{2}\!\Bigl\|\epsilon_\theta(\alpha_{t_s} \! \overline{\mathbf{z}} \!+\! \sigma_{t_s} \! \boldsymbol{\epsilon},\overline{\mathbf{c}})
\!-\!\underbrace{\frac{1}{N}\!\sum_{n=1}^{N}\!\epsilon_\theta(\alpha_{t_s} \! \mathbf{z}^n \!+\! \sigma_{t_s} \! \boldsymbol{\epsilon},\mathbf{c}^{n})}_{\text{soft target}}\Bigr\|^{2}\\
&+\frac{1}{N}\sum_{n=1}^{N} w_{t_b}\Bigr\|\epsilon_\theta(\alpha_{t_b} \! \mathbf{z}^n \!+\! \sigma_{t_b} \! \boldsymbol{\epsilon}, \mathbf{c}^{n})-\boldsymbol{\epsilon}\Bigr\|^{2} \Bigg],
\end{aligned}
\end{equation}
where the first two terms correspond to the shared phase and the last term to the branch phase.

For the \textit{shared phase}, the first term regularizes the shared stage by aligning the prediction of $\overline{\mathbf{z}}$ conditioned on $\overline{\mathbf{c}}$ with the true Gaussian noise $\boldsymbol{\epsilon}$. This ensures that the shared representation remains faithful to the original diffusion objective.
The second term enforces consistency between the model prediction conditioned on the group-level shared representation $ \overline{\mathbf{c}}$ and the averaged prediction of individual prompts in the group, which serves as a \textit{soft target}. This encourages the shared phase to capture semantic information that is representative of all prompts in the group. 
Here, $\lambda_1, \lambda_2$ balance the contributions of noise supervision and shared alignment.
For the \textit{branch phase}, the third term applies the conventional diffusion loss independently to each $(\mathbf{z}^n, \mathbf{c}^n)$ at timestep $t_b$, thereby preserving prompt-specific fidelity.  

Overall, $\mathcal{L}_{\text{SAGE}}$ balances three objectives: (i) denoising faithfulness in the shared stage, (ii) semantic alignment across prompts through soft-target distillation, and (iii) high-quality, prompt-specific generation in the branch stage.

\section{Experiments}
\label{sec:experiments}

\subsection{Experimental Setup}

\textbf{Dataset.}
To fine-tune diffusion models with SAGE, we construct a new dataset derived from MS COCO 2017 \cite{lin2014microsoft}. 
The dataset consists of over 50k prompt-image groups, each containing 2 to 5 semantically similar prompts.
Two prompts are considered semantically similar if the cosine similarity of their CLIP embeddings falls within a predefined range $(\tau_\text{min}, \tau_\text{max})$, which is tunable to simulate different real-world scenarios. The dataset is obtained by enumerating all cliques in a graph where each prompt-image pair is a node and edges connect pairs with semantically similar prompts.

\noindent \textbf{Implementation Details.}
We fine-tune Stable Diffusion v1.5 \cite{rombach2022high} on the constructed dataset using the proposed shared diffusion training algorithm.
To reduce memory and computation overhead, we adopt LoRA \cite{hu2022lora}.
The model is trained for 20k steps on a single RTX 4090 GPU with batch size of 4, using AdamW optimizer and a constant learning rate of 1e-4.

\noindent \textbf{Evaluation Metrics.}
We adopt FID \cite{heusel2017gans}, CLIP score \cite{radford2021learning}, and inter-group LPIPS \cite{zhang2018unreasonable} to evaluate image quality, text-image alignment and diversity. 
Cost saving ratio is measured as the reduction in total sampling steps required to generate the same number of images relative to independent sampling.

\subsection{Experimental Results}

Examples of generated images under shared sampling scheme for three methods---Stable Diffusion v1.5 (pre-trained), standard fine-tuning (Standard FT), and the proposed SAGE fine-tuning (SAGE FT)---are shown in Fig.~\ref{fig:images}. As can be observed, SAGE FT produces images that are more consistent with the prompts while maintaining high visual quality.

We evaluate (i) overall generation quality, (ii) robustness under varying prompt similarity, and (iii) robustness under different numbers of shared sampling steps. All experiments use DDIM sampler with 30 steps and a guidance scale of 7.5.

Table~\ref{table:overall} reports overall performance under independent and shared sampling with different sharing ratios. 
With 20\% shared steps (12.7\% cost reduction), SAGE FT achieves nearly the same performance as Standard FT under independent sampling, with only 0.2\% FID and 1.9\% CLIP gaps, showing that SAGE largely preserves quality while reducing cost. When comparing within the same sharing setting, SAGE consistently outperforms baselines---for instance, improving over Standard FT by 5.0\% in FID, 5.4\% in CLIP, and 160.0\% in diversity at 40\% sharing (25.5\% cost reduction).

\begin{table}[t]
    \vspace{-2mm}
    \centering
    \footnotesize
    \setlength{\tabcolsep}{3pt}
    \caption{Performance comparison of SD v1.5 (Pre-trained), standard fine-tuning (Standard FT) and SAGE fine-tuning (SAGE FT) under independent and shared sampling schemes with different sharing ratios $\beta=(T-T^*)/T$. Dataset parameters $\tau_\text{min}=0.6,\tau_\text{max}=0.9$ are fixed for all results.}
    \vspace{4pt}
    \begin{tabular}{cccccc}
        \toprule
        \textbf{Sampling scheme}   & \textbf{Model}   & \textbf{FID-5k$\downarrow$} & \textbf{CLIP$\uparrow$} & \textbf{Div.$\uparrow$} & \textbf{Cost saving$\uparrow$} \\
        \midrule
        \multirow{2}{*}{\makecell[c]{Independent\\sampling}} & Pre-trained & 33.41 & 0.322 & 0.697 & \multirow{2}{*}{0\%} \\
        & Standard FT & 28.38 & 0.322 & 0.698 &  \\
        \midrule
        \multirow{3}{*}{\makecell[c]{Shared sampling\\$(\beta=20\%)$}} & Pre-trained & 33.67 & 0.308 & 0.295   & \multirow{3}{*}{12.7\%} \\
        & Standard FT & 29.34 & 0.310 & 0.305 & \\
        & SAGE FT & \textbf{28.44}   & \textbf{0.316}   & \textbf{0.375}   & \\
        \midrule
        \multirow{3}{*}{\makecell[c]{Shared sampling\\$(\beta=30\%)$}} & Pre-trained & 34.13 & 0.298   & 0.190   & \multirow{3}{*}{19.1\%} \\
        & Standard FT & 29.77   & 0.302   & 0.203   & \\
        & SAGE FT & \textbf{27.71}   & \textbf{0.315}   & \textbf{0.359}   & \\
        \midrule
        \multirow{3}{*}{\makecell[c]{Shared sampling\\$(\beta=40\%)$}} & Pre-trained & 35.25 & 0.290 & 0.115 & \multirow{3}{*}{25.5\%} \\
        & Standard FT & 30.66 & 0.294 & 0.125 & \\
        & SAGE FT & \textbf{29.21}   & \textbf{0.310}   & \textbf{0.325}   & \\
        \bottomrule
    \end{tabular}
    \vspace{-3pt}
    \label{table:overall}
\end{table}

\begin{figure}[t!]\center
\includegraphics[width= 0.49 \textwidth]{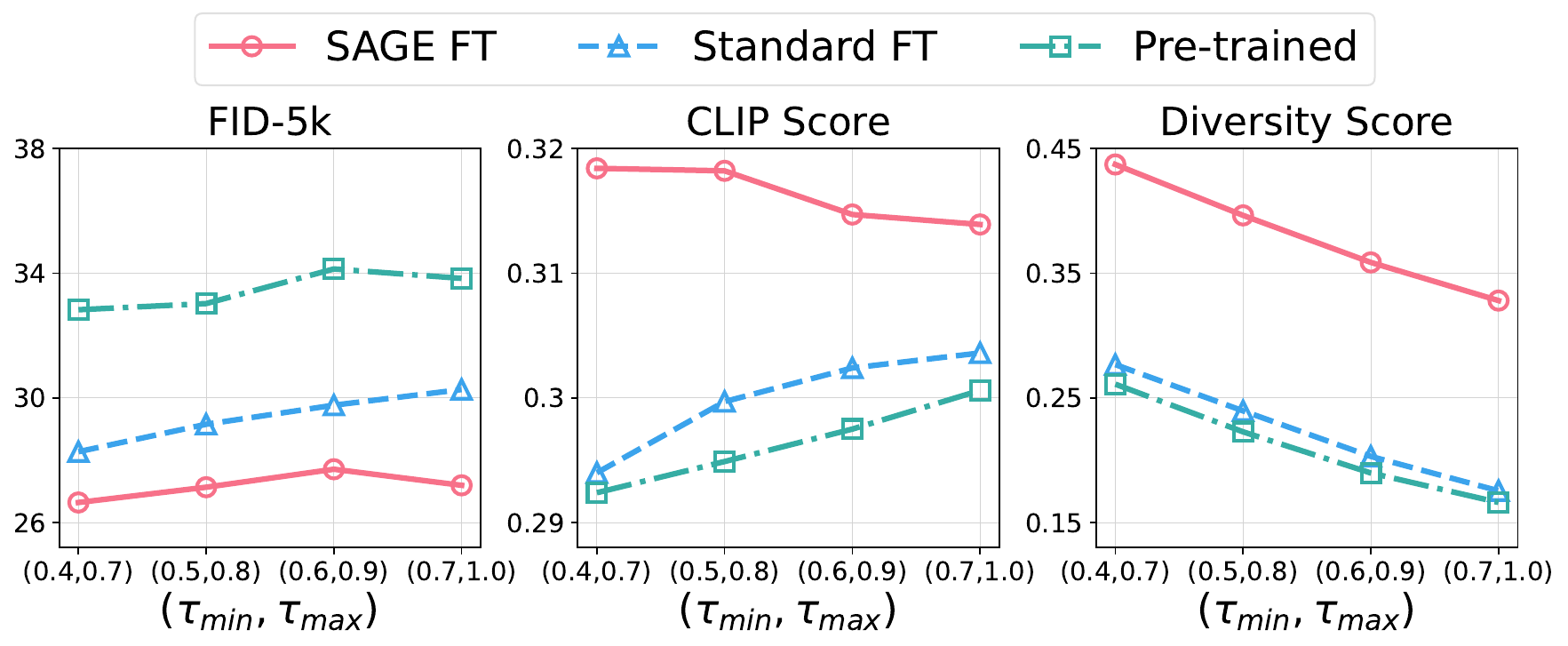}
\vspace{-18pt}
\caption{Model performance using shared sampling scheme across varying prompt similarity ranges $(\tau_{\text{min}}, \tau_{\text{max}})$ of the dataset (left$\rightarrow$right: lower$\rightarrow$higher similarity). }
\label{fig:promptsim}
\end{figure}

Fig.~\ref{fig:promptsim} further investigates the effect of prompt similarity under the shared sampling scheme. 
Across all similarity ranges, SAGE FT achieves the lowest FID, indicating superior visual quality. 
For CLIP score, SAGE FT shows a clear advantage when prompts are less similar (8.9\% higher than pre-trained), highlighting robustness in challenging scenarios and adaptability to practical use.
For diversity, while all methods degrade as similarity increases, SAGE FT maintains a consistent margin over Standard FT and pre-trained model.

\begin{figure}[t!]\center
\includegraphics[width= 0.47 \textwidth]{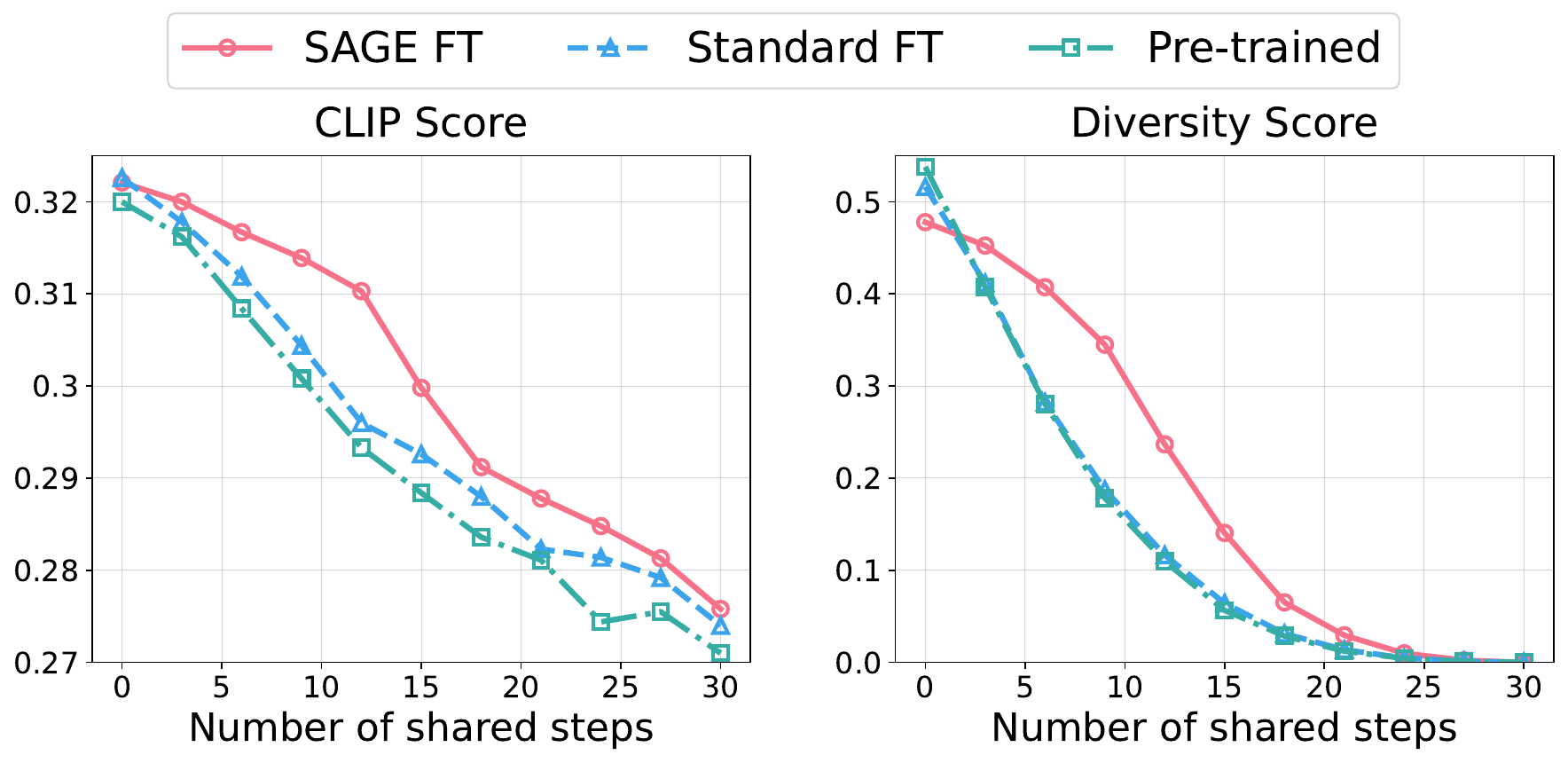}
\vspace{-6pt}
\caption{Performance of models with shared sampling scheme under varying number of shared steps. Only the sampling parameters change, training parameters are set as $\tau_\text{min}\!=\!0.6$, $\tau_\text{max}\!=\!0.9$ and $\beta\!=\!30\%$. Results are averaged over 100 prompt groups. }
\vspace{-3pt}
\label{fig:numsteps}
\end{figure}

Fig.~\ref{fig:numsteps} shows performance under varying number of shared steps. As expected, CLIP and diversity decrease with more shared steps, yet SAGE FT consistently outperforms baselines. For example, compared to Standard FT, it achieves a CLIP score 4.8\% higher at 12 shared steps, and a diversity score 118.8\% higher at 15 steps. These results demonstrate that SAGE FT can reduce computation cost while maintaining acceptable performance. FID is not reported, as it remains largely unchanged across different numbers of shared steps.

\section{Conclusion}
\label{sec:conclusion}

In this paper, we present SAGE, a framework that shares early diffusion steps across semantically similar prompts to reduce sampling cost, while enhancing generation quality under this shared sampling scheme through tailored training.
Experimental results show that SAGE reduces sampling cost while preserving high generation quality, demonstrating robustness to low prompt similarity and large sharing ratios.

\newpage
\bibliographystyle{IEEEbib}
\bibliography{refs}

\end{document}